%% file: main.tex
\documentclass[11pt,letterpaper,logo]{style}

\usepackage{algorithm}
\usepackage[noend]{algpseudocode}
\usepackage{adjustbox}
\usepackage{placeins}
\usepackage{wrapfig}
\usepackage{fontawesome5}

\title{CoWorld Reasoner: Agentic Multimodal Reasoning with the Evolving Video World Model-Driven Harness}
\runningtitle{CoWorld Reasoner}
\PublicDate{2026-10-14}
\author{%
  {\Authfont
    \textbf{Dongjie Cheng}\textsuperscript{1}\quad
    \textbf{Ruifeng Yuan}\textsuperscript{2}\quad
    \textbf{Ran Wei}\textsuperscript{2}\quad
    \textbf{Ning Song}\textsuperscript{2}\quad
    \textbf{Yongqi Li}\textsuperscript{1}\advisor\quad
    \textbf{Wenjie Li}\textsuperscript{1}
  }\\
  {\Affilfont
    \textsuperscript{1} The Hong Kong Polytechnic University\qquad
    \textsuperscript{2} Hangzhou Diagens Biotechnology Co., Ltd\\
    \texttt{dong-jie.cheng@connect.polyu.hk, liyongqi0@gmail.com}
  }
}

\begin{document}

\input{sections/abstract}
\newcommand{\TitleLinks}{%
\centering
    \vspace{6pt}
    {\noindent\absfont\fontsize{11}{13}\selectfont
    \faGithub\ Project Page: \url{https://modalitydance.github.io/CoWorld-Reasoner}\par}%
}
\maketitle

\input{sections/introduction}
\input{sections/related_work}
\input{sections/method}
\input{sections/experiments}
\setcounter{topnumber}{4}
\input{sections/conclusion}

\FloatBarrier
\begingroup
\def\UrlBreaks{\do\@\do\\\do\/\do\!\do\_\do\|\do\;\do\>\do\]\do\)\do\,\do\?\do\&\do\'\do+\do\=\do\#}
\bibliographystyle{unsrtnat}
\bibliography{references/background,references/experiments}
\endgroup

\appendix
\input{sections/appendix}

\end{document}

%% file: sections/abstract.tex
\begin{abstract}
As Vision-Language Models (VLMs) advance, they are increasingly expected to perform spatiotemporal reasoning over motion, actions, and view changes beyond what the input images show. Video World Models (VWMs) provide a natural unified engine for visual imagination by generating task-relevant visual content as additional evidence. However, existing approaches typically invoke a VWM through fixed pipelines centered on a predefined form of visual imagination, such as novel-view synthesis. To support flexible visual imagination for diverse reasoning tasks, we introduce \textbf{CoWorld Reasoner}, an agentic framework that uses a VWM as a unified visual reasoning engine. Its Video World Model-Driven Harness exposes multiple visual operations through tools and uses skills to guide the VLM in deciding whether to imagine, selecting an operation, constructing a generation request, and interpreting the resulting visual evidence. We further introduce VLM-VWM Harness Co-Evolution, which jointly revises tools, skills, and execution code based on execution feedback while keeping both model weights fixed. Across seven benchmarks spanning spatial intelligence, temporal and causal reasoning, and perceptual reasoning, CoWorld Reasoner outperforms direct reasoning with the same VLM. The strongest variant improves on all seven benchmarks, including relative accuracy gains of 11.8\% on MMSI and 13.0\% on ERQA, and outperforms all five evolving baselines using the same backbone across five benchmarks.
\end{abstract}

%% file: sections/introduction.tex
\section{Introduction}
Large Language Models (LLMs) have demonstrated strong instruction-following and reasoning abilities, but they operate primarily over text. Vision-Language Models (VLMs) extend these capabilities to visual inputs, enabling multimodal understanding and reasoning. Recent VLMs such as Qwen3.8 \citep{qwen38} and Gemini 3.8 \citep{gemini38flash} can effectively interpret images and analyze charts, and they are increasingly used in real-world multimodal applications.

Beyond static image understanding, VLMs are increasingly expected to perform spatiotemporal reasoning over motion, actions, and view changes. For example, a model may need to predict whether two objects would overlap if one were moved, reconstruct what happened between two observations, or infer what another viewpoint would reveal. Solving such problems requires going beyond observed visual evidence to reason about unobserved transitions and relations. One promising approach is visual imagination: explicitly generating task-relevant visual predictions of what is not observed and using them as additional evidence for reasoning. This raises a fundamental question: how can a VLM generate and exploit the right visual imagination for the reasoning problem at hand?

\begin{figure}[!t]
  \centering
  \includegraphics[width=\linewidth]{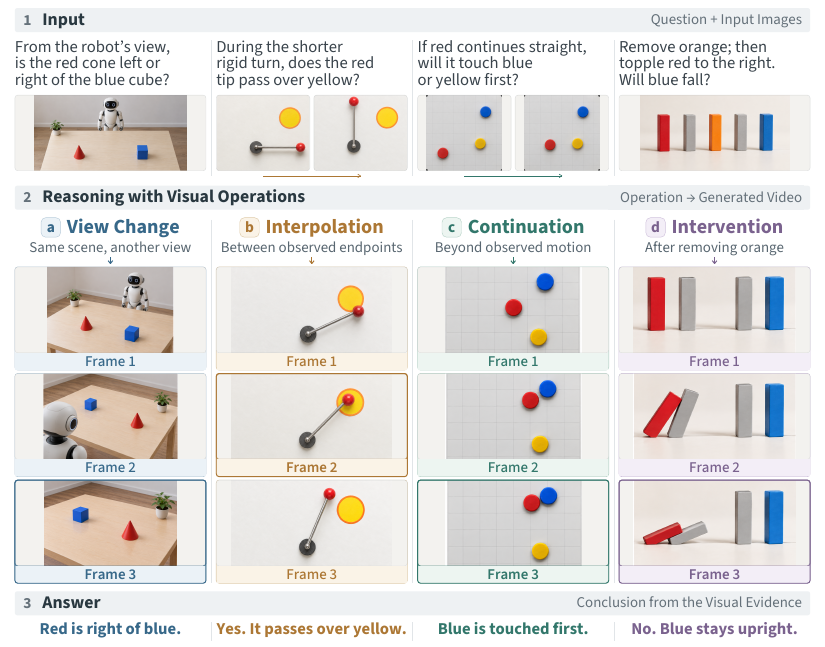}
  \caption{Illustration of how different multimodal reasoning tasks require different visual operations: (a) view change for spatial perspective taking; (b) interpolation to infer hidden motion between observed endpoints; (c) continuation to predict first contact from observed motion; and (d) intervention to reason about the outcome of a controlled change.}
  \label{fig:visual_operations}
\end{figure}

Video World Models (VWMs), such as MiniMax H3 \citep{minimaxh3} and LTX-Video \citep{ltxvideo}, provide a natural engine for visual imagination by modeling visual dynamics from large-scale video data. Given input images and an appropriate visual operation, they can generate video depicting unobserved views, intermediate states, future events, or intervention outcomes. Recent approaches, including MindJourney \citep{mindjourney} and AVIC \citep{avic}, use a VWM to generate additional visual evidence for spatial reasoning, including synthesized views for cross-view reasoning. However, these approaches typically invoke a VWM through fixed pipelines centered on a predefined form of visual imagination, such as novel-view synthesis. This limits their applicability to diverse reasoning problems, which may require fundamentally different forms of visual imagination: changing viewpoints for perspective taking, interpolating between observations to reconstruct hidden transitions, continuing observed dynamics to anticipate future events, or intervening on the scene to reason about intervention outcomes, as illustrated in Figure~\ref{fig:visual_operations}. A visual reasoner should therefore actively determine what to imagine according to the reasoning problem and how to exploit the generated visual evidence to solve it.

To enable task-dependent visual imagination, we introduce \textbf{CoWorld Reasoner}, an agentic framework that uses a VWM as a unified visual reasoning engine. Unlike an agent confined to a fixed workflow, it selects among VWM-driven tools to generate unseen views, intermediate states, continuations, or intervention outcomes according to the question. Skills guide when to invoke these tools, how to construct requests, and how to interpret the generated video; a verifier and execution code complete the harness.
A predefined harness may fail to elicit the visual evidence required by different reasoning tasks or to guide its interpretation. Refinement must coordinate evidence generation and interpretation, because what the VLM can infer depends on what the VWM can reliably generate. We therefore introduce VLM-VWM Harness Co-Evolution, which iteratively refines the interface between the VLM and VWM by jointly revising tools, skills, and execution code based on execution feedback. With both model weights fixed, co-evolution improves how the two models interact and use task-relevant visual evidence.

We evaluated CoWorld Reasoner on seven benchmarks spanning spatial intelligence \citep{mmsi,sat,spin,spatialab}, temporal and causal reasoning \citep{erqa,causalspatial}, and perceptual reasoning \citep{perceptionbench}. Across VLM/VWM variants, CoWorld Reasoner improves aggregate accuracy over direct reasoning with the same VLM. The strongest variant (GPT-5.6 and MiniMax H3) improves all seven benchmarks, with relative gains of 11.8\% on MMSI and 13.0\% on ERQA. It also outperforms all five evolving baselines using the same backbone on five benchmarks. Further analysis highlights the contributions of generated visual evidence and co-evolved control of its request and interpretation.

Our main contributions are summarized below:
\begin{enumerate}
\item We leverage a VWM as a unified visual reasoning engine for visual imagination, enabling the agent to generate task-relevant visual evidence through diverse visual operations.
\item\looseness=-1 We design a Video World Model-Driven Harness that exposes diverse visual operations through standardized tools and encodes reusable reasoning workflows as skills, enabling the agent to decide whether visual imagination is needed, which operation to invoke, and how to interpret the generated visual evidence.
\item\looseness=-1 We introduce VLM-VWM Harness Co-Evolution, which iteratively refines the interface between the VLM and VWM by jointly revising tools, skills, and execution code based on execution feedback.
\item We evaluate CoWorld Reasoner with multiple VLMs and VWMs across seven benchmarks spanning spatial intelligence, temporal and causal reasoning, and perceptual reasoning, showing gains over direct reasoning and strong performance against evolving baselines, while analyzing the contributions of the harness and its co-evolution.
\end{enumerate}

%% file: sections/related_work.tex
\section{Related Work}

\subsection{Multimodal Reasoning}
Recent work uses visual evidence to support multimodal reasoning, particularly spatial tasks.
MVoT \citep{mvot} and ThinkMorph \citep{thinkmorph} incorporate generated images into intermediate reasoning steps, while Omni-R1 \citep{omnir1generative} uses image generation to unify diverse visual operations. Thinking with Video \citep{thinkingwithvideo} uses video generation for visual reasoning. For spatial reasoning, SenseNova-SI \citep{sensenova} and OpenSpatial \citep{openspatial} use large-scale training datasets, while World2VLM \citep{world2vlm} uses views generated by world models as supervision. At inference time, MindJourney \citep{mindjourney} searches camera movements to explore a scene, and SpatialDreamer \citep{spatialdreamer} uses a learned policy to explore generated views. AVIC \citep{avic} controls world-model use and the amount of visual content generated.

\subsection{World Models}
World models capture the structure and dynamics of environments to support prediction and simulation.
Video generation models such as Wan \citep{wan}, Seedance 2.0 \citep{seedance2}, and MiniMax H3 \citep{minimaxh3} learn visual dynamics from video data, and Cosmos-Predict2.5 \citep{cosmospredict} focuses on world simulation. Marble \citep{marbleworld} generates explicit 3D environments that users can explore and edit. Interactive models such as RTFM \citep{rtfmworld} generate video in real time in response to user actions. Our approach treats a VWM as a unified visual reasoning~engine and exposes it through multiple VWM-driven tools.

\subsection{Self-Evolving Agents}

\looseness=-1
Recent approaches improve agents by iteratively revising prompts, skills, and code.
GEPA \citep{gepa} optimizes prompts from execution traces, while SkillOpt \citep{skillopt} refines reusable skills from execution feedback. DGM \citep{dgm} revises the agent's own code, while Meta-Harness \citep{metaharness} optimizes harness implementations. CoEvoSkills \citep{evoskills} jointly evolves skills and a verifier that guides their refinement. Our co-evolution supervisor jointly revises the VWM-driven harness to improve both the generation of visual evidence and the agent's interpretation of it.

%% file: sections/method.tex
\section{Method}

\begin{figure}[!t]
    \centering
    \includegraphics[width=\linewidth]{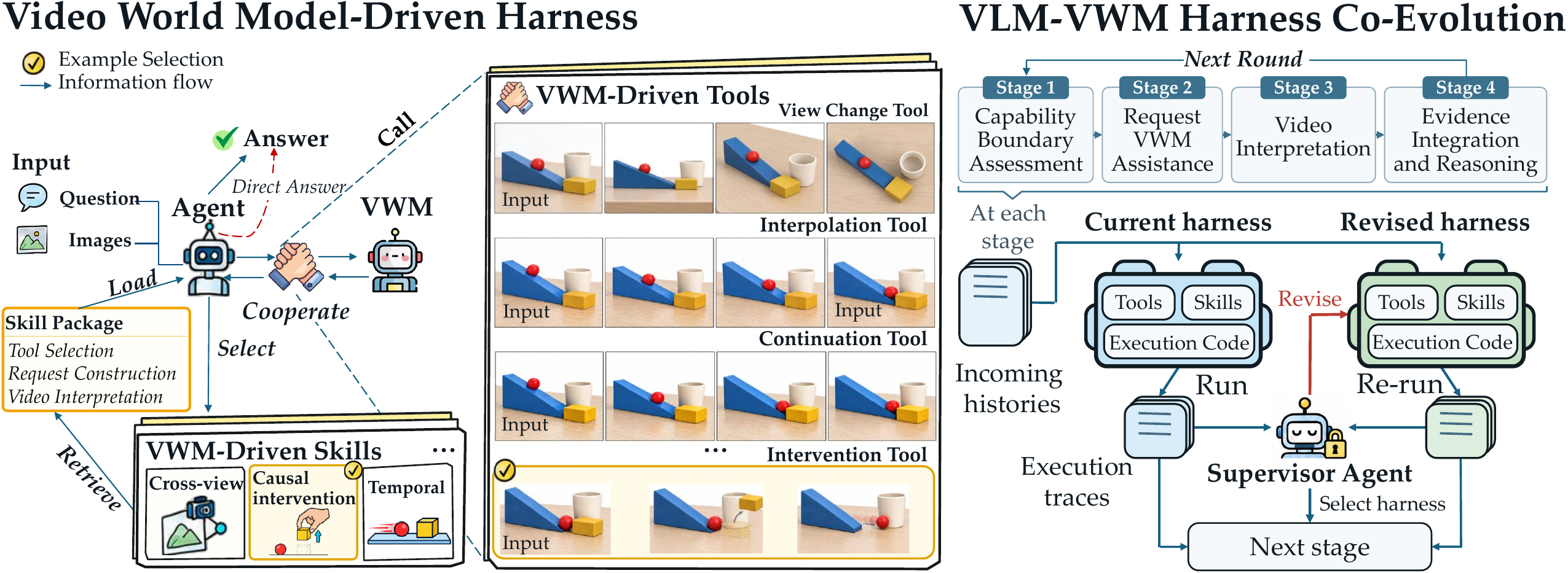}
    \caption{Overview of CoWorld Reasoner. Left: VWM-driven tools provide visual operations, while skills encode reusable workflows for the agent's tool selection, request construction, and video interpretation. Right: at each of four reasoning stages, a supervisor reviews execution traces and revises the relevant tools, skills, and execution code. The agent reruns the stage from the same incoming interaction histories. The selected harness and histories are passed to the next stage. These four stages repeat over multiple evolution rounds.}
    \label{fig:method-overview}
\end{figure}

CoWorld Reasoner uses a VWM as a unified visual reasoning engine. It has two
parts: a Video World Model-Driven Harness for requesting and interpreting
visual evidence, and VLM-VWM Harness Co-Evolution for improving the harness
through execution feedback. Figure~\ref{fig:method-overview} illustrates both
parts, and Algorithm~\ref{alg:progressive-evolution} specifies the co-evolution procedure.

We consider a question $q$ with an ordered sequence of $n$ input images
$\mathbf I=(I_1,\ldots,I_n)$. A VLM $M$ powers the agent, and a VWM $W$
provides the underlying visual generation capabilities for the VWM-driven
tools in an editable harness $H$.
The answering process can be formulated as
\begin{equation}
    \hat a=\operatorname{CoWorldReasoner}(q,\mathbf I;M,W,H),
\end{equation}
where $\operatorname{CoWorldReasoner}$ denotes harness-guided execution and $\hat a$
is the predicted answer.

\subsection{Video World Model-Driven Harness}
\label{sec:wm-reasoning}

\noindent\textbf{Harness components.} The harness comprises $L$ VWM-driven
tools $T_\ell$, $J$ VWM-driven skills $S_j$, a verifier $V$, and an
execution layer $E$:
\begin{equation}
    H=(\mathcal T,\mathcal S,V,E),\qquad
    \mathcal T=\{T_\ell\}_{\ell=1}^{L},\quad
    \mathcal S=\{S_j\}_{j=1}^{J}.
\label{eq:harness-components}
\end{equation}
Each tool $T_\ell$ exposes a visual operation through a standardized interface.
Our experiments use four types of visual operations: view change,
interpolation, continuation, and intervention.
Appendix~\ref{app:skills-tools} summarizes these operations and the corresponding
reasoning skills. A request $r$ specifies which source images to use and
in what order, the intended operation, and the conditions to preserve.
The tool executes this request through the shared VWM
and returns the generated video $\widetilde{\mathbf I}$:
\begin{equation}
    \widetilde{\mathbf I}=T_\ell(\mathbf I,r;W).
\label{eq:vwm-tool}
\end{equation}
A skill $S_j$ guides cross-view, temporal, or causal reasoning through a
reusable workflow for selecting visual operations, constructing requests,
and interpreting the generated video. For example, a cross-view skill guides
the agent to request a connecting view while preserving scene anchors, then
compare the generated video with the input images to resolve the
spatial relation. The verifier $V$ describes visible
events in the generated video and reports artifacts and uncertainty. The
execution layer $E$ contains the code that provides execution instructions, coordinates component calls,
and adds their results to the agent's interaction history.

\noindent\textbf{Agent execution.} Given $(q,\mathbf I)$, the agent assesses
whether the input images resolve the question and loads a relevant skill
when further guidance is useful. Let $h$ denote the current interaction
history and $s$ the loaded skill. The same VLM first decides on skill
loading, then on the tool and request:
\begin{equation}
    \begin{aligned}
        s &= \operatorname{Load}_{M}(q,\mathbf I,h;H),\\
        (\ell,r) &= \operatorname{Request}_{M}(q,\mathbf I,h,s;H).
    \end{aligned}
\label{eq:agent-plan}
\end{equation}
Here $s\in\mathcal S\cup\{\emptyset\}$, with $s=\emptyset$ when skill
loading is skipped. When generation is needed, the agent selects a tool index
$\ell\in\{1,\ldots,L\}$ and request $r$ to obtain the video in
Eq.~\ref{eq:vwm-tool}. If generation
is skipped, $(\ell,r)=(\emptyset,\emptyset)$ and
$\widetilde{\mathbf I}=\emptyset$.

When generated video is available, the agent compares it with the input
images. It checks whether the requested
transformation was realized and interprets the evidence relevant to $q$.
For closer inspection, it may
request a description $v$ from $V$, which analyzes only the generated video
using $q$ and $r$ as context. If verification is skipped, $v=\emptyset$.
Let $h^+$ denote the interaction history updated with the loaded guidance,
calls, and returned evidence. The agent predicts the answer $\hat a$ from
the input images and the available generated visual evidence:
\begin{equation}
    \begin{aligned}
        v &= \begin{cases}
            V(q,r,\widetilde{\mathbf I}), & \text{if verification is requested},\\
            \emptyset, & \text{otherwise}
        \end{cases}\\
        \hat a &= M\!\left(q,\mathbf I,\widetilde{\mathbf I},v;h^+\right).
    \end{aligned}
\label{eq:evidence-integration}
\end{equation}

\subsection{VLM-VWM Harness Co-Evolution}
\label{sec:harness-evolution}

VLM-VWM Harness Co-Evolution jointly refines the components that govern visual
generation and evidence use, while keeping the VLM and VWM weights fixed.
These components are coupled: changes to request construction can alter the
visual cues available in the generated video, which in turn affect the guidance
needed to interpret and integrate them.

Agent execution follows four reasoning stages. At each stage, co-evolution
alternates between forward execution and feedback-guided revision. The agent
first runs the current harness on the development set. A supervisor powered by the same VLM $M$ uses the resulting
feedback to revise the harness and rerun that stage from the same incoming
interaction histories. This cycle can repeat before the supervisor selects the best
candidate and passes its harness and histories to the next stage.
We repeat these four stages for $R$ evolution rounds.

\begin{algorithm}[!t]
\small
\caption{VLM-VWM Harness Co-Evolution.}
\label{alg:progressive-evolution}
\begin{algorithmic}[1]
\Require VLM: $M$; VWM: $W$; Initial harness: $H_1^{(0)}$; Development set: $\mathcal D_{\mathrm{dev}}$;
\newline Evolution rounds: $R$; Revision budget per stage: $B$
\For{$t=1,\ldots,R$}
    \State $\tau_0^*\gets\Call{InitHistories}{\mathcal D_{\mathrm{dev}}}$
    \For{$k=1,\ldots,4$}
        \State $\mathcal C_k\gets\emptyset$
        \For{$e=0,\ldots,B$}
            \State $\tau_k^{(e)}\gets\Call{RunStage}{\mathcal D_{\mathrm{dev}},H_k^{(e)},\tau_{k-1}^*,k}$
            \State $P_k^{(e)}\gets\Call{Summarize}{\tau_k^{(e)},H_k^{(e)},k;\text{available scorer feedback}}$
            \State Add $(H_k^{(e)},\tau_k^{(e)},P_k^{(e)})$ to $\mathcal C_k$
            \If{$e=B$} \textbf{break}
            \EndIf
            \State $\Delta_k^{(e)}\gets\operatorname{Review}_{M}(H_k^{(e)},P_k^{(e)};k)$
            \If{the supervisor ends revision} \textbf{break}
            \EndIf
            \State $H_k^{(e+1)}\gets\Call{Revise}{H_k^{(e)},\Delta_k^{(e)}}$
        \EndFor
        \State $(H_k^*,\tau_k^*)\gets\operatorname{Select}_{M}(\mathcal C_k;k)$
        \If{$k<4$} $H_{k+1}^{(0)}\gets H_k^*$
        \EndIf
    \EndFor
    \If{$t<R$} $H_1^{(0)}\gets H_4^*$
    \EndIf
\EndFor
\State \Return $H_4^*$
\end{algorithmic}
\end{algorithm}

\noindent\textbf{Progressive execution.}
During \emph{Capability Boundary Assessment} ($k=1$), the agent judges whether
the VWM can supply useful missing evidence and may load a relevant skill.
In \emph{Request VWM Assistance} ($k=2$), it can revise this assessment before
constructing and executing a generation request or skipping generation.
During \emph{Video Interpretation} ($k=3$), it reviews the input
images and interprets any generated video alongside them, optionally
requesting verification.
Finally, in \emph{Evidence Integration and Reasoning} ($k=4$), it combines
the input images with generated visual evidence to infer the final answer.
The stage and candidate notation below is local to the current round.
At stage $k$, each candidate $H_k^{(e)}$
after $e$ revisions is run from the same selected interaction histories $\tau_{k-1}^*$,
retaining earlier reasoning and evidence. Execution on the development set
$\mathcal D_{\mathrm{dev}}$ produces updated histories $\tau_k^{(e)}$:
\begin{equation}
    \tau_k^{(e)}
    =\operatorname{RunStage}\!\left(\mathcal D_{\mathrm{dev}},
      H_k^{(e)},\tau_{k-1}^*,k\right).
\label{eq:progressive-evidence}
\end{equation}
Here $e=0$ denotes the stage's initial candidate, and $\tau_0^*$ contains
the initialized task histories.

\noindent\textbf{Feedback-guided revision and selection.} Each run produces stage
feedback $P_k^{(e)}$: stage-specific metrics, sampled execution traces, and, when
an answer is available, correctness feedback from an external answer scorer.
During capability boundary assessment, for example, the supervisor examines
skill-loading rates and sampled load/skip decisions to refine guidance on
when and which skills to load.
The supervisor reads
additional execution traces, generated video, and harness
source files as needed to link recurring problems to the relevant harness
components.

The proposed changes $\Delta_k^{(e)}$ can affect tool interfaces,
skill workflows, verifier and stage instructions, and execution code:
\begin{equation}
    \begin{aligned}
        \Delta_k^{(e)}
        &=\operatorname{Review}_{M}\!\left(H_k^{(e)},P_k^{(e)};k\right),\\
        H_k^{(e+1)}
        &=\operatorname{Revise}\!\left(H_k^{(e)},\Delta_k^{(e)}\right).
    \end{aligned}
\label{eq:harness-revision}
\end{equation}
Each revised candidate is evaluated via Eq.~\ref{eq:progressive-evidence}.
Appendix~\ref{app:stage-assessment} summarizes the checks used to assess
stages and select candidates.

Each stage allows up
to $B$ revisions, as specified in Appendix~\ref{app:implementation}; the supervisor may stop earlier. Let
$\mathcal C_k$ collect the initial candidate and all completed revisions,
each stored with its histories and stage feedback. The supervisor
selects the best completed candidate by comparing stage-specific metrics and sampled
execution traces, weighing targeted improvements against regressions across tasks.
This selection returns the harness $H_k^*$ and histories $\tau_k^*$:
\begin{equation}
    \begin{aligned}
        (H_k^*,\tau_k^*)
        &=\operatorname{Select}_{M}(\mathcal C_k;k),\\
        H_{k+1}^{(0)}&=H_k^*,\qquad k<4.
    \end{aligned}
\label{eq:candidate-handoff}
\end{equation}
At each hand-off, the next stage resumes from the selected harness and its
corresponding histories. After Stage~4, the selected harness $H_4^*$
initializes Stage~1 of the next round, with task histories reinitialized
from the original questions and images. Later revisions can thus affect
earlier-stage behavior in subsequent rounds. After $R$ rounds, the final
selected harness is used for inference.

%% file: sections/experiments.tex
\section{Experiments}
\setcounter{topnumber}{1}
\subsection{Experimental Setup}

\paragraph{Benchmarks.}

We evaluated CoWorld Reasoner on the seven benchmarks in Table~\ref{tab:main_results}. MMSI \citep{mmsi} assesses reasoning about positions and motion across images; SAT \citep{sat} covers ego and object motion, action outcomes, goals, and perspective taking; and Spin \citep{spin} tests perspective taking under translation and rotation. SpatiaLab \citep{spatialab} uses Multiple-Choice (MCQ) and Open-Ended (Open) questions about real scenes. ERQA \citep{erqa} combines spatial, physical, and world knowledge, while CausalSpatial \citep{causalspatial} tests reasoning about collisions, occlusions, and trajectories under object interventions. PerceptionBench \citep{perceptionbench} covers counting, attributes, and text reading.

\paragraph{Baselines.}

We compare with five evolving agents adapted to the same GPT-5.6 inputs and scoring protocol (HSI, ReasoningBank, Meta-Harness, ExpeL, and GEPA) and use direct GPT-5.6 reasoning as the reference. The comparison tables also include published results for fine-tuned embodied models, specialist spatial models, and vision-based agents; mechanisms, configurations, and source-protocol notes are provided in Appendix~\ref{app:experimental_details}.

\paragraph{Evaluation Protocol.}

We use 120 development examples, excluded from testing, to evolve CoWorld Reasoner and the five evolving baselines; 30 examples come from each of MMSI, Spin, CausalSpatial, and PerceptionBench. Accuracy is $100\times N_{\mathrm{correct}}/N$; differences are in percentage points, and averages are item-weighted. We measure error recovery as accuracy on the subset of test items that the GPT-5.6 direct reasoning baseline answers incorrectly. Appendix~\ref{app:experimental_details} details evaluation sizes and scoring protocols.

\paragraph{Implementation.}

We used GPT-5.6 for the task agent, co-evolution supervisor, and external answer scorer, and MiniMax H3 Max Turbo \citep{minimaxh3} as the VWM. We tested the final co-evolved harness without further updates. Appendix~\ref{app:implementation} specifies model and generation settings; Appendix~\ref{app:stage-assessment} summarizes candidate assessment.

\begin{table}[t]
\centering
\caption{Multimodal reasoning accuracy (\%, $\uparrow$). Base Models and Fine-tuned Models use published results; dashes denote benchmark scores unreported in these sources. Avg. requires complete benchmark coverage. SL and PB denote SpatiaLab and PerceptionBench. }
\label{tab:main_results}
\begingroup
\definecolor{sectiongray}{HTML}{F0F2F3}
\definecolor{oursbg}{HTML}{E8F2F1}
\definecolor{accent}{HTML}{19796F}
\definecolor{missing}{HTML}{C4C8CC}
\small
\setlength{\tabcolsep}{1.6pt}
\renewcommand{\arraystretch}{0.94}
\resizebox{\linewidth}{!}{%
\begin{tabular}{@{}>{\raggedright\arraybackslash}p{150pt}*{3}{>{\centering\arraybackslash}p{30pt}}*{2}{>{\centering\arraybackslash}p{39pt}}*{2}{>{\centering\arraybackslash}p{40pt}}>{\centering\arraybackslash}p{55pt}>{\centering\arraybackslash}p{32pt}@{}}
\toprule
& \multicolumn{5}{c}{\textbf{Spatial Intelligence}} & \multicolumn{2}{c}{\textbf{Temporal/Causal}} & \textbf{Perception} & \\
\cmidrule(lr){2-6}\cmidrule(lr){7-8}\cmidrule(lr){9-9}
\textbf{Method} & \makebox[29pt][c]{\textbf{MMSI}} & \textbf{SAT} & \textbf{Spin} & \makebox[38pt][c]{\textbf{SL-MCQ}} & \makebox[38pt][c]{\textbf{SL-Open}} & \textbf{ERQA} & \makebox[29pt][c]{\textbf{Causal}} & \textbf{PB} & \textbf{Avg.} \\
\midrule
\rowcolor{sectiongray}[0pt][0pt]\multicolumn{10}{@{}l@{}}{\strut\textit{Base Models}} \\
GPT-5-mini & 34.10 & 75.40 & \textcolor{missing}{\textemdash} & 54.29 & \textbf{40.93} & \textcolor{missing}{\textemdash} & 45.44 & \textcolor{missing}{\textemdash} & \textcolor{missing}{\textemdash} \\
GPT-5 & 41.90 & 80.10 & 78.50 & \textcolor{missing}{\textemdash} & \textcolor{missing}{\textemdash} & \textcolor{missing}{\textemdash} & 54.17 & \textcolor{missing}{\textemdash} & \textcolor{missing}{\textemdash} \\
Gemini-2.5-Pro & 36.90 & 76.70 & 75.20 & 50.07 & 33.61 & 55.80 & 50.55 & \textcolor{missing}{\textemdash} & \textcolor{missing}{\textemdash} \\
Qwen3-VL-8B & 31.10 & 52.70 & 67.40 & \textcolor{missing}{\textemdash} & \textcolor{missing}{\textemdash} & 40.10 & 43.53 & \textcolor{missing}{\textemdash} & \textcolor{missing}{\textemdash} \\
\addlinespace[3pt]
\rowcolor{sectiongray}[0pt][0pt]\multicolumn{10}{@{}l@{}}{\strut\textit{Fine-tuned Models}} \\
MiMo-Embodied & 30.20 & 78.67 & \textcolor{missing}{\textemdash} & \textcolor{missing}{\textemdash} & \textcolor{missing}{\textemdash} & 46.75 & \textcolor{missing}{\textemdash} & \textcolor{missing}{\textemdash} & \textcolor{missing}{\textemdash} \\
RoboBrain2.0 & 28.50 & 86.67 & \textcolor{missing}{\textemdash} & \textcolor{missing}{\textemdash} & \textcolor{missing}{\textemdash} & 46.00 & \textcolor{missing}{\textemdash} & \textcolor{missing}{\textemdash} & \textcolor{missing}{\textemdash} \\
\addlinespace[3pt]
\rowcolor{sectiongray}[0pt][0pt]\multicolumn{10}{@{}l@{}}{\strut\textit{Evolving Agents}} \\
GPT-5.6 & 38.35 & 88.33 & 82.24 & 56.64 & 36.21 & 51.75 & 60.72 & 50.84 & 58.29 \\
\textit{+~HSI} & 35.05 & 87.67 & 73.46 & 55.71 & 29.57 & 57.25 & 58.06 & 48.92 & 54.39 \\
\textit{+~ReasoningBank} & 36.80 & 86.67 & 81.69 & 56.57 & 30.29 & 59.00 & 60.65 & 50.07 & 57.30 \\
\textit{+~Meta-Harness} & 37.42 & 88.33 & 83.31 & 55.86 & 30.57 & \textbf{60.00} & 60.07 & 49.83 & 57.63 \\
\textit{+~ExpeL} & 37.42 & 89.00 & 82.36 & 56.64 & 29.57 & 57.00 & 61.22 & 49.12 & 57.25 \\
\textit{+~GEPA} & 38.87 & 87.67 & 79.92 & 54.86 & 28.21 & 59.75 & \textbf{62.51} & 47.61 & 56.25 \\
\rowcolor{oursbg}\mbox{\textbf{+ CoWorld Reasoner}} & \textbf{42.89} & \textbf{89.67} & \textbf{83.76} & \textbf{57.36} & 37.64 & 58.50 & 60.79 & \textbf{51.35} & \textbf{59.69} \\
\textit{Gain (pp)} & \textcolor{accent}{+4.54} & \textcolor{accent}{+1.33} & \textcolor{accent}{+1.51} & \textcolor{accent}{+0.71} & \textcolor{accent}{+1.43} & \textcolor{accent}{+6.75} & \textcolor{accent}{+0.07} & \textcolor{accent}{+0.51} & \textcolor{accent}{+1.40} \\
\addlinespace[3pt]
\bottomrule
\end{tabular}
}
\endgroup
\end{table}

\begin{table}[t]
\centering
\caption{Spatial intelligence accuracy on MMSI and SAT (\%, $\uparrow$). PR, Mot., MSR, and Persp. denote positional relationship, motion, multi-step reasoning, and perspective taking, respectively. Dashes denote scores unreported in the source publications. Best results are bolded.}
\label{tab:spatial_specialists}
\begingroup
\definecolor{sectiongray}{HTML}{F0F2F3}
\definecolor{oursbg}{HTML}{E8F2F1}
\definecolor{accent}{HTML}{19796F}
\definecolor{missing}{HTML}{C4C8CC}
\small
\setlength{\tabcolsep}{1.6pt}
\renewcommand{\arraystretch}{1.18}
\resizebox{\linewidth}{!}{%
\begin{tabular}{@{}>{\raggedright\arraybackslash}p{147pt}*{5}{>{\centering\arraybackslash}p{29pt}}>{\centering\arraybackslash}p{32pt}*{5}{>{\centering\arraybackslash}p{29pt}}@{}}
\toprule
& \multicolumn{5}{c}{\textbf{MMSI}} & \multicolumn{6}{c}{\textbf{SAT}} \\
\cmidrule(lr){2-6}\cmidrule(l){7-12}
\textbf{Method} & \textbf{PR} & \textbf{Attr.} & \textbf{Mot.} & \textbf{MSR} & \textbf{All} & \textbf{Ego} & \textbf{Obj} & \textbf{Act.} & \textbf{Goal} & \textbf{Persp.} & \textbf{All} \\
\midrule
\rowcolor{sectiongray}[0pt][0pt]\multicolumn{12}{@{}l@{}}{\strut\textit{Base Models}} \\
o1 & \textcolor{missing}{\textemdash} & \textcolor{missing}{\textemdash} & \textcolor{missing}{\textemdash} & \textcolor{missing}{\textemdash} & \textcolor{missing}{\textemdash} & 78.30 & 82.60 & 73.00 & 73.50 & 69.70 & 74.60 \\
GPT-4.1 & 30.65 & 30.77 & 34.67 & 28.79 & 30.90 & 95.70 & 73.90 & 78.30 & 88.20 & 39.40 & 74.00 \\
GPT-5.4 & \textcolor{missing}{\textemdash} & \textcolor{missing}{\textemdash} & \textcolor{missing}{\textemdash} & \textcolor{missing}{\textemdash} & 37.50 & \textcolor{missing}{\textemdash} & \textcolor{missing}{\textemdash} & \textcolor{missing}{\textemdash} & \textcolor{missing}{\textemdash} & \textcolor{missing}{\textemdash} & \textcolor{missing}{\textemdash} \\
Qwen3-VL-8B & 33.72 & 26.15 & 28.00 & 29.80 & 31.10 & \textcolor{missing}{\textemdash} & \textcolor{missing}{\textemdash} & \textcolor{missing}{\textemdash} & \textcolor{missing}{\textemdash} & \textcolor{missing}{\textemdash} & 52.70 \\
\addlinespace[2pt]
\rowcolor{sectiongray}[0pt][0pt]\multicolumn{12}{@{}l@{}}{\strut\textit{Fine-tuned Models}} \\
SenseNova-SI & \textbf{52.87} & 40.77 & 32.67 & 27.78 & \textbf{43.30} & \textcolor{missing}{\textemdash} & \textcolor{missing}{\textemdash} & \textcolor{missing}{\textemdash} & \textcolor{missing}{\textemdash} & \textcolor{missing}{\textemdash} & \textcolor{missing}{\textemdash} \\
ViewFusion & 39.66 & 35.38 & 36.67 & 23.23 & 35.40 & \textcolor{missing}{\textemdash} & \textcolor{missing}{\textemdash} & \textcolor{missing}{\textemdash} & \textcolor{missing}{\textemdash} & \textcolor{missing}{\textemdash} & \textcolor{missing}{\textemdash} \\
OpenSpatial & \textcolor{missing}{\textemdash} & \textcolor{missing}{\textemdash} & \textcolor{missing}{\textemdash} & \textcolor{missing}{\textemdash} & 41.90 & \textcolor{missing}{\textemdash} & \textcolor{missing}{\textemdash} & \textcolor{missing}{\textemdash} & \textcolor{missing}{\textemdash} & \textcolor{missing}{\textemdash} & \textcolor{missing}{\textemdash} \\
World2VLM & \textcolor{missing}{\textemdash} & \textcolor{missing}{\textemdash} & \textcolor{missing}{\textemdash} & \textcolor{missing}{\textemdash} & \textcolor{missing}{\textemdash} & \textbf{100.00} & 65.22 & 67.57 & 79.41 & 57.57 & 72.67 \\
MiMo-Embodied & \textcolor{missing}{\textemdash} & \textcolor{missing}{\textemdash} & \textcolor{missing}{\textemdash} & \textcolor{missing}{\textemdash} & 30.20 & \textcolor{missing}{\textemdash} & \textcolor{missing}{\textemdash} & \textcolor{missing}{\textemdash} & \textcolor{missing}{\textemdash} & \textcolor{missing}{\textemdash} & 78.67 \\
RoboBrain2.0 & 26.44 & 33.85 & 30.00 & 29.29 & 28.50 & \textcolor{missing}{\textemdash} & \textcolor{missing}{\textemdash} & \textcolor{missing}{\textemdash} & \textcolor{missing}{\textemdash} & \textcolor{missing}{\textemdash} & 86.67 \\
Spatial-Interactor & \textcolor{missing}{\textemdash} & \textcolor{missing}{\textemdash} & \textcolor{missing}{\textemdash} & \textcolor{missing}{\textemdash} & 32.20 & \textcolor{missing}{\textemdash} & \textcolor{missing}{\textemdash} & \textcolor{missing}{\textemdash} & \textcolor{missing}{\textemdash} & \textcolor{missing}{\textemdash} & 74.00 \\
\addlinespace[2pt]
\rowcolor{sectiongray}[0pt][0pt]\multicolumn{12}{@{}l@{}}{\strut\textit{Agents with Vision Models}} \\
AVIC & 31.23 & 44.62 & 35.33 & 32.32 & 33.80 & \textbf{100.00} & 78.20 & 83.70 & 85.20 & 54.50 & 79.30 \\
AVIC-R & \textcolor{missing}{\textemdash} & \textcolor{missing}{\textemdash} & \textcolor{missing}{\textemdash} & \textcolor{missing}{\textemdash} & \textcolor{missing}{\textemdash} & 91.30 & 86.90 & 83.70 & 85.20 & 57.50 & 80.00 \\
MindJourney & \textcolor{missing}{\textemdash} & \textcolor{missing}{\textemdash} & \textcolor{missing}{\textemdash} & \textcolor{missing}{\textemdash} & \textcolor{missing}{\textemdash} & 95.70 & 82.60 & 83.80 & 88.20 & 75.80 & 84.70 \\
SandboxVLM & \textcolor{missing}{\textemdash} & \textcolor{missing}{\textemdash} & \textcolor{missing}{\textemdash} & \textcolor{missing}{\textemdash} & \textcolor{missing}{\textemdash} & \textbf{100.00} & 82.60 & 79.40 & \textbf{92.90} & 70.00 & 84.10 \\
\addlinespace[2pt]
\rowcolor{sectiongray}[0pt][0pt]\multicolumn{12}{@{}l@{}}{\strut\textit{Evolving Agents}} \\
GPT-5.6 & 35.59 & 42.40 & 40.00 & 41.62 & 38.35 & \textbf{100.00} & 95.65 & 89.19 & 83.82 & 78.79 & 88.33 \\
\textit{+~HSI} & 30.22 & 42.40 & 38.62 & 40.10 & 35.05 & \textbf{100.00} & 91.30 & 91.89 & 82.35 & 77.27 & 87.67 \\
\textit{+~ReasoningBank} & 33.60 & 43.20 & 40.69 & 38.07 & 36.80 & 97.83 & 91.30 & 87.84 & 83.82 & 77.27 & 86.67 \\
\textit{+~Meta-Harness} & 34.19 & 45.60 & 36.55 & 41.12 & 37.42 & 91.30 & \textbf{97.83} & \textbf{93.24} & 80.88 & 81.82 & 88.33 \\
\textit{+~ExpeL} & 32.41 & \textbf{48.80} & 37.24 & \textbf{43.15} & 37.42 & \textbf{100.00} & \textbf{97.83} & 89.19 & 83.82 & 80.30 & 89.00 \\
\textit{+~GEPA} & 36.18 & 44.00 & 39.31 & 42.13 & 38.87 & 93.48 & 93.48 & 89.19 & 85.29 & 80.30 & 87.67 \\
\rowcolor{oursbg}\mbox{\textbf{+ CoWorld Reasoner}} & 42.15 & 46.40 & \textbf{43.45} & 42.13 & 42.89 & \textbf{100.00} & 93.48 & 89.19 & 85.29 & \textbf{84.85} & \textbf{89.67} \\
\textit{Gain (pp)} & \textcolor{accent}{+6.56} & \textcolor{accent}{+4.00} & \textcolor{accent}{+3.45} & \textcolor{accent}{+0.51} & \textcolor{accent}{+4.54} & \textcolor{accent}{+0.00} & \textcolor{gray}{-2.17} & \textcolor{accent}{+0.00} & \textcolor{accent}{+1.47} & \textcolor{accent}{+6.06} & \textcolor{accent}{+1.33} \\
\bottomrule
\end{tabular}
}
\endgroup
\end{table}

\subsection{Main Results}

\paragraph{Overall Performance.}
Table~\ref{tab:main_results} shows that CoWorld Reasoner, using its final co-evolved harness, improves over the GPT-5.6 direct reasoning baseline on all seven benchmarks, especially on MMSI and ERQA, with gains of 4.54 and 6.75 percentage points, respectively. It leads all five evolving baselines on five benchmarks and achieves the highest aggregate accuracy. All five evolving baselines also score below the GPT-5.6 direct reasoning baseline overall. The advantage over the best evolving baseline reaches 7.07 points on open-ended SpatiaLab. Gains also extend to SAT, SpatiaLab, and ERQA, which provide no development examples; these results show that the final harness transfers across benchmarks. Appendix~\ref{app:evolving-analysis} further compares development and test performance.


\paragraph{Spatial Intelligence.}
As shown in Table~\ref{tab:spatial_specialists}, gains span all four MMSI categories and reach 6.56 points for positional relationships. Overall accuracy approaches SenseNova-SI's reported result. SAT perspective taking reaches 84.85\%, the highest among the compared methods; goal reasoning also improves, while object motion declines. Accuracy on Spin likewise improves in the View, Mental Rotation, and Perspective categories, as detailed in Appendix~\ref{app:subset-results}. The cross-view skill guides the agent to track object identity and interpret relations in the question's reference frame.

\subsection{Analysis}

\begin{figure}[t]
\centering
\includegraphics[width=\linewidth]{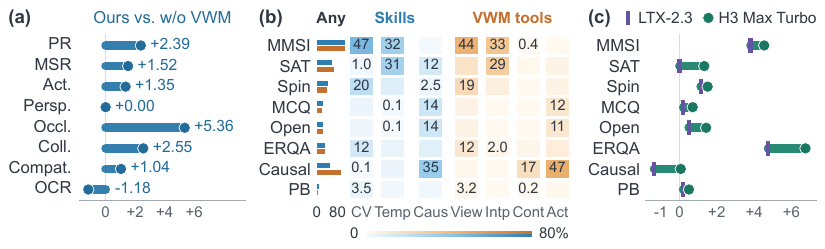}
\caption{Analysis of VWM contributions. (a) Subtask gains over Ours (w/o VWM). (b) Skill-loading and tool-use rates by benchmark; tool calls are grouped by visual operation (\% of items). (c) Benchmark gains over the GPT-5.6 direct reasoning baseline with two VWMs.}
\label{fig:vwm_analysis}
\end{figure}

\paragraph{Contribution of VWM-Driven Tools.}
To assess the contribution of generated visual evidence, we disabled VWM-driven tools and the verifier while retaining the evolved skills and four-stage execution. Figure~\ref{fig:vwm_analysis}a shows the largest tool gains on CausalSpatial Occlusion (Occl.) and Collision (Coll.). Generated video exposes the intermediate interactions needed to answer these questions. SAT Perspective Taking (Persp.) already improves without tools and gains no further benefit from generation. Gains without tools indicate that harness guidance can also improve reasoning \mbox{over the input images}.

\paragraph{VWM-Driven Skill and Tool Use.}
We measured skill-loading and tool-use rates to examine task-dependent selection. Figure~\ref{fig:vwm_analysis}b shows that on MMSI, the agent mainly loads cross-view (CV) and temporal (Temp) skills and calls view change (View) and interpolation (Intp) tools to relate views and infer hidden motion. On CausalSpatial, it uses the causal intervention (Caus) skill with intervention (Act) and continuation (Cont) tools to predict intervention outcomes. Skills and tools are rarely used on PerceptionBench, where direct perception often suffices.

\paragraph{Effect of VWM Choice.}
We compared H3 Max Turbo with LTX-2.3 using GPT-5.6, retaining the evolved skills and four-stage execution. Both VWM variants improve overall accuracy and performance on MMSI, Spin, and SpatiaLab, as shown in Figure~\ref{fig:vwm_analysis}c. The H3 variant yields larger gains across the evaluated benchmarks and also improves SAT, while LTX-2.3 falls below GPT-5.6 on CausalSpatial. These results show that the harness can reuse spatial reasoning workflows across VWMs and benefit from a stronger visual generation engine.

\Needspace{14\baselineskip}
\begingroup
\setlength{\intextsep}{4pt}
\setlength{\columnsep}{12pt}
\begin{wrapfigure}[11]{r}{0.42\linewidth}
\vspace{-6pt}
\centering
\includegraphics[width=\linewidth]{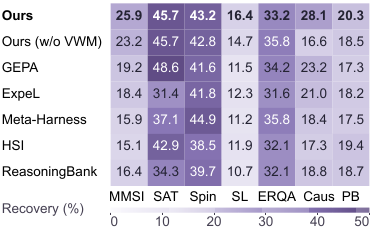}
\captionsetup{width=\linewidth}
\caption{Recovery on GPT-5.6 errors; SL combines SpatiaLab MCQ and Open.}
\label{fig:recovery}
\end{wrapfigure}
\paragraph{Recovery from GPT-5.6 Errors.}
For each benchmark, we evaluated all harness variants and evolving baselines on the subset of test examples that the GPT-5.6 direct reasoning baseline answered incorrectly. CoWorld Reasoner leads on MMSI, SpatiaLab, CausalSpatial, and PerceptionBench (Figure~\ref{fig:recovery}). Without VWM tools, it exceeds the evolving baselines on MMSI and SpatiaLab; adding them improves CausalSpatial recovery, leaves SAT unchanged, and reduces ERQA recovery. These CausalSpatial gains mirror the occlusion and collision improvements in Figure~\ref{fig:vwm_analysis}a: the harness resolves additional GPT-5.6 errors while its overall CausalSpatial accuracy remains close to the GPT-5.6 baseline.

\par\endgroup

\begingroup
\setlength{\intextsep}{4pt}
\setlength{\columnsep}{12pt}
\begin{wrapfigure}[11]{r}{0.42\linewidth}
\vspace{-6pt}
\centering
\includegraphics[width=\linewidth]{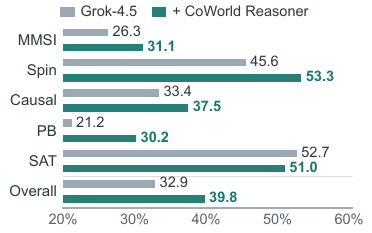}
\captionsetup{width=\linewidth}
\caption{Grok-4.5 accuracy with direct reasoning and the transferred harness.}
\label{fig:harness_transfer}
\end{wrapfigure}
\paragraph{Harness Transfer Across VLMs.}
We transferred the GPT-5.6 harness to Grok-4.5 and ran one additional co-evolution round before evaluation. Figure~\ref{fig:harness_transfer} shows that item-weighted accuracy rises from 32.86\% to 39.78\%, with the largest gains on PerceptionBench and Spin. MMSI and CausalSpatial also improve, while SAT declines slightly. The gains span tasks that use the input images directly and tasks that require reasoning about unseen views or outcomes. After adaptation, Grok-4.5 reuses the source harness's procedures for selecting, interpreting, and integrating visual evidence across benchmarks.

\par\endgroup

%% file: sections/conclusion.tex
\section{Conclusion and Future Work}
We presented CoWorld Reasoner, which treats a VWM as a unified visual reasoning engine through a Video World Model-Driven Harness of tools, skills, and a verifier. VLM-VWM Harness Co-Evolution uses execution feedback to improve how the agent requests, evaluates, and integrates visual evidence. CoWorld Reasoner improves over direct reasoning with the same VLM on all seven benchmarks and outperforms all five evolving baselines on five of these benchmarks. Agentic tool use and harness co-evolution extend VWM-based reasoning beyond fixed novel-view synthesis pipelines. Future work will explore test-time training to co-evolve the harness, VLM, and VWM for visual generation and multimodal reasoning across diverse reasoning tasks.

%% file: sections/appendix.tex
\section{Appendix}\label{app:experimental_details}

\subsection{Evaluation Data}

Table~\ref{tab:appendix_data_sizes} summarizes the development and evaluation sets. We used the 120 development examples to evolve the reusable components of our method. We adapted the five evolving baselines to the same visual inputs and scoring interface, then evaluated the selected variants on benchmarks excluding these development examples.

\begin{table}[htbp]
\centering
\caption{\centering Development examples and evaluation items.}
\label{tab:appendix_data_sizes}
\begingroup
\small
\setlength{\tabcolsep}{3pt}
\renewcommand{\arraystretch}{1.12}
\begin{tabular*}{\linewidth}{@{\extracolsep{\fill}}lcccccccc@{}}
\toprule
 & \textbf{MMSI} & \textbf{SAT} & \textbf{Spin} & \multicolumn{2}{c}{\textbf{SpatiaLab}} & \textbf{ERQA} & \textbf{CausalSpatial} & \textbf{PerceptionBench} \\
\cmidrule(lr){5-6}
 & & & & \textbf{MCQ} & \textbf{Open} & & & \\
\midrule
\textbf{Development} & 30 & 0 & 30 & 0 & 0 & 0 & 30 & 30 \\
\textbf{Evaluation} & 970 & 300 & 2,709 & 1,400 & 1,400 & 400 & 1,395 & 2,970 \\
\bottomrule
\end{tabular*}
\par\smallskip
\begin{minipage}{\linewidth}
\small
SAT evaluates 150 questions under two option permutations, yielding 300 items. SpatiaLab MCQ and Open use the same 1,400 questions in two answer formats.
\end{minipage}
\endgroup
\end{table}

\paragraph{Baseline configurations.}
\looseness=-1
The evolving-agent comparison uses HSI \citep{hsi}, ReasoningBank \citep{reasoningbank}, Meta-Harness \citep{metaharness}, ExpeL \citep{expel}, and GEPA \citep{gepa}, all adapted to GPT-5.6 with the same visual inputs and scoring interface. Table~\ref{tab:evolving-baselines} summarizes their main mechanisms. They operate directly on the input images and evolve prompts, memories, or code; none uses VWM-driven tools. Our harness co-evolves VWM-driven tools, skills, and execution code. The direct GPT-5.6 system answers from the input images and is the reference baseline.

For the published comparison rows, Qwen3-VL~\citep{qwen3vl} uses the 8B-Instruct model; MiMo-Embodied~\citep{mimoembodied} and RoboBrain2.0~\citep{robobrain2} use their 7B and 32B models, respectively. Spatial-Interactor~\citep{spatialinteractor} uses Qwen3-VL-8B with SFT and on-policy distillation. SandboxVLM~\citep{sandboxvlm} uses GPT-5-mini; AVIC-R uses a Qwen2.5-VL-7B policy with GPT-4.1 for question answering. Table~\ref{tab:spatial_specialists} also includes ViewFusion~\citep{viewfusion}. MMSI category scores for Qwen3-VL and RoboBrain2.0 are computed by weighting reported subcategory scores by sample count \citep{sensenova,mmsi}. MiMo-Embodied's MMSI score and RoboBrain2.0's ERQA score are reported by \citet{rynnbrain11}. Avg. is reported for methods evaluated on all benchmarks and weighted by the evaluation sizes in Table~\ref{tab:appendix_data_sizes}.

\begin{table}[t]
\centering
\caption{Main mechanisms of the evolving-agent baselines.}
\label{tab:evolving-baselines}
\begingroup
\small
\setlength{\tabcolsep}{5pt}
\renewcommand{\arraystretch}{1.18}
\begin{tabularx}{\linewidth}{@{}>{\raggedright\arraybackslash}p{88pt}>{\raggedright\arraybackslash}X@{}}
\toprule
\textbf{Method} & \textbf{Main mechanism} \\
\midrule
HSI & Hierarchically revises task-specific agent harnesses. \\
\addlinespace[2pt]
ReasoningBank & Builds and retrieves reusable reasoning memories from agent experience. \\
\addlinespace[2pt]
Meta-Harness & Proposes and evaluates changes to executable harness code. \\
\addlinespace[2pt]
ExpeL & Distills insights from task experience and retrieves examples for later tasks. \\
\addlinespace[2pt]
GEPA & Reflectively revises prompts using execution and evaluation feedback. \\
\bottomrule
\end{tabularx}
\endgroup
\end{table}

\subsection{Scoring and Usage Statistics}

We computed accuracy over all items in each benchmark or category. We normalized multiple-choice answers before comparing them with the correct option. For PerceptionBench, we used deterministic option matching when an answer maps to a supplied choice; otherwise, a GPT-5.6 judge evaluates correctness under the benchmark's general-QA criterion. The same judge evaluates semantic equivalence for open-ended SpatiaLab questions. The judge operates in an evaluation context fully independent of the task agent.

Recovery measures accuracy on the subset of test items answered incorrectly by the GPT-5.6 direct reasoning baseline. When combining SpatiaLab MCQ and Open, we aggregated correct answers and item counts \mbox{across both formats}.

Skill-loading rate is the percentage of items on which the corresponding skill is loaded. Tool-use rate is the percentage of items with a VWM-driven tool call for the corresponding visual operation. An item can contribute to multiple columns; the Any column counts items with at least one skill load or tool call in the corresponding group.

\begin{figure}[!t]
\centering
\includegraphics[width=\linewidth]{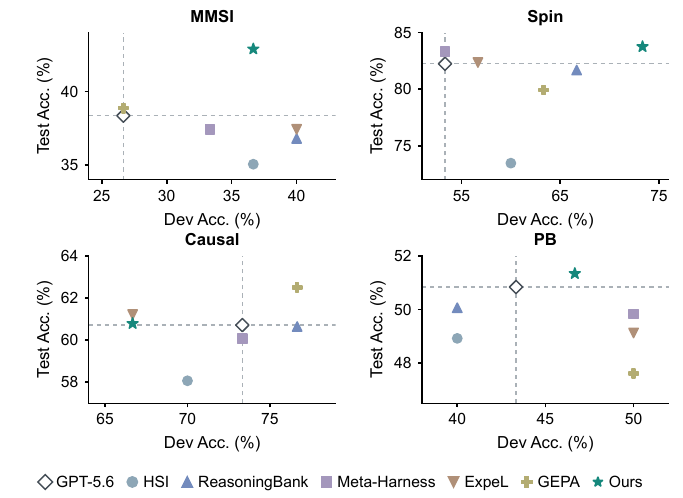}
\caption{\centering Development and test accuracy. Dashed lines show the GPT-5.6 direct reasoning baseline; stars mark CoWorld Reasoner.}
\label{fig:development_test_profile}
\end{figure}

\subsection{Agent and Generation Configuration}
\label{app:implementation}

The task agent, co-evolution supervisor, and external answer scorer each use the \mbox{gpt-5.6-luna} deployment in separate contexts. The GPT-5.6 direct reasoning baseline answers each item from the input images in one stage. All five evolving baselines use this GPT-5.6 deployment. Table~\ref{tab:appendix_configuration} lists agent and \mbox{video generation settings}.

The agent carries its interaction history through stages 1--4, following the order in Section~\ref{sec:harness-evolution}. The verifier receives the generated video. The ``Ours (w/o VWM)'' variant retains the VWM-driven skills and four-stage execution while disabling all VWM-driven tools and the verifier.

We used the MiniMax H3 Max Turbo image-to-video endpoint on FAL. A request supplies one source image or two images defining the start and end frames. Generation requests describe the visual transformation without prescribing the task answer.

\begin{table}[htbp]
\centering
\caption{\centering Agent and video generation settings.}
\label{tab:appendix_configuration}
\begingroup
\small
\definecolor{configgroup}{HTML}{F0F2F3}
\setlength{\tabcolsep}{4pt}
\renewcommand{\arraystretch}{1.18}
\begin{tabularx}{\linewidth}{@{}>{\raggedright\arraybackslash}X>{\raggedleft\arraybackslash}p{56pt}@{}p{16pt}@{}>{\raggedright\arraybackslash}X>{\raggedleft\arraybackslash}p{56pt}@{}}
\toprule
\multicolumn{2}{@{}>{\columncolor{configgroup}[0pt][0pt]}l@{}}{\strut\textbf{Task agent}} & & \multicolumn{2}{@{}>{\columncolor{configgroup}[0pt][0pt]}l@{}}{\strut\textbf{Video generation}} \\
\midrule
Reasoning effort & Medium & & Duration & 5 seconds \\
Temperature & 0 & & Resolution & 480P \\
Max. output tokens/model call & 4,096 & & Prompt expansion & Balanced \\
Model-turn limits (stages 1--4) & 4 / 6 / 6 / 6 & & Frames for agent reasoning & 9 \\
\bottomrule
\end{tabularx}
\endgroup
\end{table}

We used $R=3$ evolution rounds. At each stage, the supervisor evaluates the initial candidate and up to $B=5$ revisions, with early stopping.

For transfer to Grok-4.5, we initialized the harness from the completed GPT-5.6 co-evolution, replaced the VLM with Grok-4.5, and ran one additional co-evolution round before evaluation.

\subsection{VWM-Driven Skills and Visual Operations}
\label{app:skills-tools}

Table~\ref{tab:skills-tools} summarizes the three skills and four visual operation categories, using the abbreviations in Figure~\ref{fig:vwm_analysis}b.

\begin{table}[htbp]
\centering
\caption{Skills and visual operations used in the experiments.}
\label{tab:skills-tools}
\begingroup
\small
\definecolor{harnessgroup}{HTML}{F0F2F3}
\setlength{\tabcolsep}{5pt}
\renewcommand{\arraystretch}{1.20}
\begin{tabularx}{\linewidth}{@{}>{\raggedright\arraybackslash}p{112pt}>{\centering\arraybackslash}p{26pt}>{\raggedright\arraybackslash}X@{}}
\toprule
\textbf{Component} & \textbf{Abbr.} & \textbf{Function} \\
\midrule
\rowcolor{harnessgroup}[0pt][0pt]\multicolumn{3}{@{}l@{}}{\strut\textit{VWM-driven skills}} \\
\addlinespace[2pt]
Cross-view & CV & Guides reasoning about identity, viewpoint, and \mbox{layout relations across images}. \\
\addlinespace[2pt]
Temporal & Temp & Guides reasoning about transitions between ordered observations and about short temporal continuations. \\
\addlinespace[2pt]
Causal intervention & Caus & Guides reasoning about the outcome of a specified action \mbox{on a visible target}. \\
\addlinespace[4pt]
\rowcolor{harnessgroup}[0pt][0pt]\multicolumn{3}{@{}l@{}}{\strut\textit{VWM-driven visual operations}} \\
\addlinespace[2pt]
View change & View & Generates a new or connecting viewpoint to expose \mbox{a spatial relation}. \\
\addlinespace[2pt]
Interpolation & Intp & Generates intermediate states between ordered observations. \\
\addlinespace[2pt]
Continuation & Cont & Extends an observed state forward or backward in time. \\
\addlinespace[2pt]
Intervention & Act & Generates the predicted outcome of a specified action \mbox{on a visible target}. \\
\bottomrule
\end{tabularx}
\endgroup
\end{table}

\subsection{Stage Assessment and Candidate Selection}
\label{app:stage-assessment}

At each evolution round, the supervisor selects the best completed candidate using stage-specific assessments, sampled traces, and answer-correctness feedback. It checks skill choice, generated visual evidence coverage, request validity, source-image order, tool execution, interpreted events and artifacts, reference frames, and answer correctness across the four stages. It weighs cross-benchmark gains against regressions.

\subsection{Development and Test Performance}
\label{app:evolving-analysis}

Figure~\ref{fig:development_test_profile} compares the final evolved agents on the 120 development examples and 8,044 held-out test examples from MMSI, Spin, CausalSpatial, and PerceptionBench. Aggregate accuracy is weighted by the number of examples within each split.

ReasoningBank improves over GPT-5.6 on MMSI, Spin, and CausalSpatial during development, yet falls below it on all three test sets. Across the four benchmarks, CoWorld Reasoner and \mbox{ReasoningBank} both reach an aggregate development accuracy of 55.83\%, but their test accuracies are 62.88\% and 60.95\%, respectively, compared with 61.62\% for GPT-5.6. CoWorld Reasoner improves over GPT-5.6 on each of the four test benchmarks.

\input{sections/appendix_subset_tables}

\subsection{Results by Task Category}
\label{app:subset-results}

Tables~\ref{tab:appendix_spin_subsets}--\ref{tab:appendix_pb_subsets} report category accuracies for Spin, SpatiaLab, ERQA, CausalSpatial, and PerceptionBench. All denotes overall benchmark accuracy, and Gain (pp) is the difference from GPT-5.6 in percentage points. Best results are bolded. Category results for MMSI and SAT appear in Table~\ref{tab:spatial_specialists}.

\subsection{Examples of Evolved Guidance}
\label{app:evolved-guidance}

\paragraph{Binding requests to their visual sources.}
The request-construction instructions align image references with the source order supplied to the VWM. In Figure~\ref{fig:case-study-panel}, the second input image becomes Picture~1 in the generation request. They also ground object-relative movement in visible front and rear features. The interpretation instructions guide the agent to check object identity and express spatial relations in the reference frame specified by the question.

\paragraph{Distinguishing view changes from object motion.}
\looseness=-1
The interpretation instructions guide the agent and verifier to compare the subject with background landmarks across early, middle, and late frames. A turning subject against a fixed background can indicate object rotation rather than camera motion. In Figure~\ref{fig:case-study-motion}, the agent tracks the SUV relative to the stationary chair, plant, and vase, then relates its motion to the input image and question.

Figures~\ref{fig:case-study-panel}--\ref{fig:case-study-intervention} show reasoning trajectories for six examples. In Figure~\ref{fig:case-study-removal}, the generated video shows the remaining plates staying in place after the bottom plate is removed, prompting the agent to revise its initial answer. Boldface highlights key cues, and \texttt{[...]} indicates omitted skill text.

\input{sections/appendix_case_studies}

%% file: sections/appendix_subset_tables.tex
\begin{table}[!htbp]
\centering
\makebox[\linewidth][l]{%
\begin{minipage}{\linewidth}
\caption{Spin accuracy by task category (\%, $\uparrow$).}
\label{tab:appendix_spin_subsets}
\begingroup
\definecolor{sectiongray}{HTML}{F0F2F3}
\definecolor{oursbg}{HTML}{E8F2F1}
\definecolor{accent}{HTML}{19796F}
\small
\setlength{\tabcolsep}{1.3pt}
\renewcommand{\arraystretch}{1.15}
\begin{tabularx}{\linewidth}{@{}>{\raggedright\arraybackslash}p{90pt}>{\hsize=0.87\hsize\linewidth=\hsize\centering\arraybackslash}X>{\hsize=0.95\hsize\linewidth=\hsize\centering\arraybackslash}X>{\hsize=1.3\hsize\linewidth=\hsize\centering\arraybackslash}X>{\hsize=0.95\hsize\linewidth=\hsize\centering\arraybackslash}X>{\hsize=0.67\hsize\linewidth=\hsize\centering\arraybackslash}X>{\hsize=1.28\hsize\linewidth=\hsize\centering\arraybackslash}X>{\hsize=1.3\hsize\linewidth=\hsize\centering\arraybackslash}X>{\hsize=0.68\hsize\linewidth=\hsize\centering\arraybackslash}X@{}}
\toprule
{\textbf{Method}} & {\textbf{Identity}} & {\textbf{Relation}} & {\textbf{Translation}} & {\textbf{Rotation}} & {\textbf{View}} & {\textbf{\mbox{Mental rot.}}} & {\textbf{Perspective}} & {\textbf{All}} \\
\midrule
\rowcolor{sectiongray}[0pt][0pt]\multicolumn{9}{@{}l@{}}{\strut\textit{Evolving Agents}} \\
GPT-5.6 & 98.51 & 88.68 & \textbf{100.00} & \textbf{90.65} & 67.17 & 66.97 & 69.00 & 82.24 \\
\textit{+~HSI} & 99.01 & 64.78 & 94.87 & 82.72 & 66.26 & 64.68 & 61.83 & 73.46 \\
\textit{+~ReasoningBank} & \textbf{99.26} & 84.43 & \textbf{100.00} & \textbf{90.65} & 68.09 & 72.48 & 68.03 & 81.69 \\
\textit{+~Meta-Harness} & 98.76 & 88.52 & 98.72 & 89.24 & 65.65 & \textbf{76.61} & \textbf{72.27} & 83.31 \\
\textit{+~ExpeL} & 98.76 & 87.74 & \textbf{100.00} & 86.12 & 66.87 & 71.56 & 71.45 & 82.36 \\
\textit{+~GEPA} & 98.51 & 78.14 & 98.08 & 88.10 & 65.96 & 72.48 & 70.31 & 79.92 \\
\rowcolor{oursbg}\textbf{+ Ours} & 99.01 & \textbf{91.35} & \textbf{100.00} & 87.25 & \textbf{71.73} & 69.72 & 71.13 & \textbf{83.76} \\
\textit{Gain (pp)} & \textcolor{accent}{+0.50} & \textcolor{accent}{+2.67} & \textcolor{accent}{+0.00} & \textcolor{gray}{-3.40} & \textcolor{accent}{+4.56} & \textcolor{accent}{+2.75} & \textcolor{accent}{+2.12} & \textcolor{accent}{+1.51} \\
\bottomrule
\end{tabularx}%
\endgroup
\end{minipage}%
}
\end{table}

\begin{table}[!htbp]
\centering
\makebox[\linewidth][l]{%
\begin{minipage}{\linewidth}
\caption{SpatiaLab (MCQ) accuracy by task category (\%, $\uparrow$).}
\label{tab:appendix_spatialab_mcq_subsets}
\begingroup
\definecolor{sectiongray}{HTML}{F0F2F3}
\definecolor{oursbg}{HTML}{E8F2F1}
\definecolor{accent}{HTML}{19796F}
\small
\setlength{\tabcolsep}{2pt}
\renewcommand{\arraystretch}{1.15}
\begin{tabularx}{\linewidth}{@{}>{\raggedright\arraybackslash}p{105pt}>{\hsize=0.9\hsize\linewidth=\hsize\centering\arraybackslash}X>{\hsize=1.2\hsize\linewidth=\hsize\centering\arraybackslash}X>{\hsize=0.9\hsize\linewidth=\hsize\centering\arraybackslash}X>{\hsize=1\hsize\linewidth=\hsize\centering\arraybackslash}X>{\hsize=1.03\hsize\linewidth=\hsize\centering\arraybackslash}X>{\hsize=1.2\hsize\linewidth=\hsize\centering\arraybackslash}X>{\hsize=0.77\hsize\linewidth=\hsize\centering\arraybackslash}X@{}}
\toprule
{\textbf{Method}} & {\textbf{3D geo.}} & {\textbf{Depth/occl.}} & {\textbf{Orient.}} & {\textbf{Rel. pos.}} & {\textbf{Size/scale}} & {\textbf{Navigation}} & {\textbf{All}} \\
\midrule
\rowcolor{sectiongray}[0pt][0pt]\multicolumn{8}{@{}l@{}}{\strut\textit{Evolving Agents}} \\
GPT-5.6 & 47.90 & 55.21 & 60.40 & \textbf{67.92} & 53.17 & 57.38 & 56.64 \\
\textit{+~HSI} & 47.48 & 57.53 & 60.40 & 64.62 & 50.40 & 55.70 & 55.71 \\
\textit{+~ReasoningBank} & 47.48 & 59.46 & 60.40 & 63.68 & \textbf{54.37} & 55.27 & 56.57 \\
\textit{+~Meta-Harness} & 47.06 & 57.53 & 59.41 & 66.04 & 49.60 & 57.38 & 55.86 \\
\textit{+~ExpeL} & 49.58 & \textbf{59.85} & 60.89 & 67.45 & 52.78 & 51.05 & 56.64 \\
\textit{+~GEPA} & 49.58 & 55.98 & 59.90 & 61.32 & 46.83 & 57.38 & 54.86 \\
\rowcolor{oursbg}\textbf{+ Ours} & \textbf{52.10} & 57.14 & \textbf{64.36} & 67.45 & 48.02 & \textbf{57.81} & \textbf{57.36} \\
\textit{Gain (pp)} & \textcolor{accent}{+4.20} & \textcolor{accent}{+1.93} & \textcolor{accent}{+3.96} & \textcolor{gray}{-0.47} & \textcolor{gray}{-5.16} & \textcolor{accent}{+0.42} & \textcolor{accent}{+0.71} \\
\bottomrule
\end{tabularx}%
\endgroup
\end{minipage}%
}
\end{table}

\begin{table}[!htbp]
\centering
\makebox[\linewidth][l]{%
\begin{minipage}{\linewidth}
\caption{SpatiaLab (Open) accuracy by task category (\%, $\uparrow$).}
\label{tab:appendix_spatialab_open_subsets}
\begingroup
\definecolor{sectiongray}{HTML}{F0F2F3}
\definecolor{oursbg}{HTML}{E8F2F1}
\definecolor{accent}{HTML}{19796F}
\small
\setlength{\tabcolsep}{2pt}
\renewcommand{\arraystretch}{1.15}
\begin{tabularx}{\linewidth}{@{}>{\raggedright\arraybackslash}p{105pt}>{\hsize=0.9\hsize\linewidth=\hsize\centering\arraybackslash}X>{\hsize=1.2\hsize\linewidth=\hsize\centering\arraybackslash}X>{\hsize=0.9\hsize\linewidth=\hsize\centering\arraybackslash}X>{\hsize=1\hsize\linewidth=\hsize\centering\arraybackslash}X>{\hsize=1.03\hsize\linewidth=\hsize\centering\arraybackslash}X>{\hsize=1.2\hsize\linewidth=\hsize\centering\arraybackslash}X>{\hsize=0.77\hsize\linewidth=\hsize\centering\arraybackslash}X@{}}
\toprule
{\textbf{Method}} & {\textbf{3D geo.}} & {\textbf{Depth/occl.}} & {\textbf{Orient.}} & {\textbf{Rel. pos.}} & {\textbf{Size/scale}} & {\textbf{Navigation}} & {\textbf{All}} \\
\midrule
\rowcolor{sectiongray}[0pt][0pt]\multicolumn{8}{@{}l@{}}{\strut\textit{Evolving Agents}} \\
GPT-5.6 & 34.87 & 28.19 & 36.14 & \textbf{41.51} & 40.48 & 37.13 & 36.21 \\
\textit{+~HSI} & 26.89 & 23.55 & 30.69 & 37.26 & 30.56 & 29.96 & 29.57 \\
\textit{+~ReasoningBank} & 27.73 & 23.55 & 32.18 & 33.96 & 34.13 & 31.22 & 30.29 \\
\textit{+~Meta-Harness} & 28.57 & 23.55 & 33.17 & 37.74 & 32.94 & 29.11 & 30.57 \\
\textit{+~ExpeL} & 28.15 & 22.39 & 29.70 & 34.43 & 30.95 & 32.91 & 29.57 \\
\textit{+~GEPA} & 25.63 & 23.55 & 32.67 & 31.60 & 27.38 & 29.96 & 28.21 \\
\rowcolor{oursbg}\textbf{+ Ours} & \textbf{37.82} & \textbf{30.12} & \textbf{38.12} & 41.04 & \textbf{40.87} & \textbf{38.82} & \textbf{37.64} \\
\textit{Gain (pp)} & \textcolor{accent}{+2.94} & \textcolor{accent}{+1.93} & \textcolor{accent}{+1.98} & \textcolor{gray}{-0.47} & \textcolor{accent}{+0.40} & \textcolor{accent}{+1.69} & \textcolor{accent}{+1.43} \\
\bottomrule
\end{tabularx}%
\endgroup
\end{minipage}%
}
\end{table}

\begin{table}[!htbp]
\centering
\makebox[\linewidth][l]{%
\begin{minipage}{\linewidth}
\caption{ERQA accuracy by task category (\%, $\uparrow$).}
\label{tab:appendix_erqa_subsets}
\begingroup
\definecolor{sectiongray}{HTML}{F0F2F3}
\definecolor{oursbg}{HTML}{E8F2F1}
\definecolor{accent}{HTML}{19796F}
\small
\setlength{\tabcolsep}{1.5pt}
\renewcommand{\arraystretch}{1.15}
\begin{tabularx}{\linewidth}{@{}>{\raggedright\arraybackslash}p{90pt}>{\hsize=0.98\hsize\linewidth=\hsize\centering\arraybackslash}X>{\hsize=1.36\hsize\linewidth=\hsize\centering\arraybackslash}X>{\hsize=0.86\hsize\linewidth=\hsize\centering\arraybackslash}X>{\hsize=1.15\hsize\linewidth=\hsize\centering\arraybackslash}X>{\hsize=1\hsize\linewidth=\hsize\centering\arraybackslash}X>{\hsize=0.79\hsize\linewidth=\hsize\centering\arraybackslash}X>{\hsize=0.76\hsize\linewidth=\hsize\centering\arraybackslash}X>{\hsize=1.35\hsize\linewidth=\hsize\centering\arraybackslash}X>{\hsize=0.75\hsize\linewidth=\hsize\centering\arraybackslash}X@{}}
\toprule
{\textbf{Method}} & {\textbf{Action}} & {\textbf{\mbox{Multi-view}}} & {\textbf{Other}} & {\textbf{Pointing}} & {\textbf{Spatial}} & {\textbf{State}} & {\textbf{Task}} & {\textbf{Trajectory}} & {\textbf{All}} \\
\midrule
\rowcolor{sectiongray}[0pt][0pt]\multicolumn{10}{@{}l@{}}{\strut\textit{Evolving Agents}} \\
GPT-5.6 & 45.83 & 35.14 & 21.43 & 79.41 & 57.14 & 47.27 & \textbf{65.79} & 48.48 & 51.75 \\
\textit{+~HSI} & 54.17 & 40.54 & 42.86 & 79.41 & 59.52 & 56.36 & \textbf{65.79} & 54.55 & 57.25 \\
\textit{+~ReasoningBank} & 48.61 & \textbf{48.65} & \textbf{57.14} & 79.41 & 60.71 & 65.45 & 63.16 & 56.06 & 59.00 \\
\textit{+~Meta-Harness} & 52.78 & 43.24 & \textbf{57.14} & \textbf{85.29} & 57.14 & \textbf{70.91} & \textbf{65.79} & 56.06 & \textbf{60.00} \\
\textit{+~ExpeL} & 50.00 & 43.24 & 42.86 & 79.41 & 59.52 & 63.64 & 60.53 & 53.03 & 57.00 \\
\textit{+~GEPA} & 54.17 & \textbf{48.65} & 42.86 & 73.53 & \textbf{63.10} & 63.64 & 57.89 & \textbf{62.12} & 59.75 \\
\rowcolor{oursbg}\textbf{+ Ours} & \textbf{56.94} & \textbf{48.65} & 35.71 & \textbf{85.29} & 58.33 & 52.73 & 60.53 & 60.61 & 58.50 \\
\textit{Gain (pp)} & \textcolor{accent}{+11.11} & \textcolor{accent}{+13.51} & \textcolor{accent}{+14.29} & \textcolor{accent}{+5.88} & \textcolor{accent}{+1.19} & \textcolor{accent}{+5.45} & \textcolor{gray}{-5.26} & \textcolor{accent}{+12.12} & \textcolor{accent}{+6.75} \\
\bottomrule
\end{tabularx}%
\endgroup
\end{minipage}%
}
\end{table}

\begin{table}[!htbp]

\centering
\makebox[\linewidth][c]{%
\begin{minipage}{0.88\linewidth}
\caption{CausalSpatial accuracy by task category (\%, $\uparrow$).}
\label{tab:appendix_causal_subsets}
\begingroup
\definecolor{sectiongray}{HTML}{F0F2F3}
\definecolor{oursbg}{HTML}{E8F2F1}
\definecolor{accent}{HTML}{19796F}
\normalsize
\setlength{\tabcolsep}{2pt}
\renewcommand{\arraystretch}{1.15}
\begin{tabularx}{\linewidth}{@{}>{\raggedright\arraybackslash}p{105pt}>{\hsize=1\hsize\linewidth=\hsize\centering\arraybackslash}X>{\hsize=1\hsize\linewidth=\hsize\centering\arraybackslash}X>{\hsize=1\hsize\linewidth=\hsize\centering\arraybackslash}X>{\hsize=1\hsize\linewidth=\hsize\centering\arraybackslash}X>{\hsize=1\hsize\linewidth=\hsize\centering\arraybackslash}X@{}}
\toprule
{\textbf{Method}} & {\textbf{Collision}} & {\textbf{Compat.}} & {\textbf{Occlusion}} & {\textbf{Physics}} & {\textbf{All}} \\
\midrule
\rowcolor{sectiongray}[0pt][0pt]\multicolumn{6}{@{}l@{}}{\strut\textit{Evolving Agents}} \\
GPT-5.6 & 54.92 & \textbf{64.58} & 57.74 & 76.62 & 60.72 \\
\textit{+~HSI} & 52.13 & 60.42 & 51.79 & 76.62 & 58.06 \\
\textit{+~ReasoningBank} & 53.95 & \textbf{64.58} & 58.93 & 78.25 & 60.65 \\
\textit{+~Meta-Harness} & 54.68 & 59.38 & 58.93 & 75.32 & 60.07 \\
\textit{+~ExpeL} & 53.83 & \textbf{64.58} & 63.10 & 78.90 & 61.22 \\
\textit{+~GEPA} & \textbf{55.65} & 59.38 & \textbf{64.29} & 80.84 & \textbf{62.51} \\
\rowcolor{oursbg}\textbf{+ Ours} & 54.56 & 62.50 & 52.98 & \textbf{81.17} & 60.79 \\
\textit{Gain (pp)} & \textcolor{gray}{-0.36} & \textcolor{gray}{-2.08} & \textcolor{gray}{-4.76} & \textcolor{accent}{+4.55} & \textcolor{accent}{+0.07} \\
\bottomrule
\end{tabularx}%
\endgroup
\end{minipage}%
}
\end{table}

\begin{table}[!htbp]

\centering
\makebox[\linewidth][l]{%
\begin{minipage}{\linewidth}
\caption{PerceptionBench accuracy by error category (\%, $\uparrow$).}
\label{tab:appendix_pb_subsets}
\begingroup
\definecolor{sectiongray}{HTML}{F0F2F3}
\definecolor{oursbg}{HTML}{E8F2F1}
\definecolor{accent}{HTML}{19796F}
\fontsize{9.5}{11.5}\selectfont
\setlength{\tabcolsep}{1pt}
\renewcommand{\arraystretch}{1.15}
\begin{tabularx}{\linewidth}{@{}>{\raggedright\arraybackslash}p{95pt}>{\hsize=1.25\hsize\linewidth=\hsize\centering\arraybackslash}X>{\hsize=1.35\hsize\linewidth=\hsize\centering\arraybackslash}X>{\hsize=1.13\hsize\linewidth=\hsize\centering\arraybackslash}X>{\hsize=1.05\hsize\linewidth=\hsize\centering\arraybackslash}X>{\hsize=0.85\hsize\linewidth=\hsize\centering\arraybackslash}X>{\hsize=0.85\hsize\linewidth=\hsize\centering\arraybackslash}X>{\hsize=0.94\hsize\linewidth=\hsize\centering\arraybackslash}X>{\hsize=1\hsize\linewidth=\hsize\centering\arraybackslash}X>{\hsize=0.85\hsize\linewidth=\hsize\centering\arraybackslash}X>{\hsize=0.85\hsize\linewidth=\hsize\centering\arraybackslash}X>{\hsize=0.88\hsize\linewidth=\hsize\centering\arraybackslash}X@{}}
\toprule
{\scriptsize\textbf{Method}} & {\scriptsize\textbf{Context}} & {\scriptsize\textbf{\mbox{Depth/3D}}} & {\scriptsize\textbf{\mbox{Fine gr.}}} & {\scriptsize\textbf{Halluc.}} & {\scriptsize\textbf{OCR}} & {\scriptsize\textbf{Attr.}} & {\scriptsize\textbf{Comp.}} & {\scriptsize\textbf{Count}} & {\scriptsize\textbf{Loc.}} & {\scriptsize\textbf{Rel.}} & {\scriptsize\textbf{All}} \\
\midrule
\rowcolor{sectiongray}[0pt][0pt]\multicolumn{12}{@{}l@{}}{\strut\textit{Evolving Agents}} \\
GPT-5.6 & 46.03 & 44.95 & \textbf{48.60} & 26.49 & 50.98 & 53.37 & 56.36 & \textbf{49.24} & 65.44 & \textbf{62.08} & 50.84 \\
\textit{+~HSI} & 41.67 & 43.12 & 46.15 & 27.24 & 49.02 & 50.00 & 51.64 & \textbf{49.24} & \textbf{66.06} & 59.63 & 48.92 \\
\textit{+~ReasoningBank} & 47.22 & 42.20 & 45.10 & 29.10 & \textbf{52.16} & 54.29 & \textbf{57.82} & 47.40 & 63.30 & 58.72 & 50.07 \\
\textit{+~Meta-Harness} & 44.84 & 43.73 & 46.15 & 26.49 & 51.76 & 53.68 & 57.45 & 47.40 & 62.08 & 60.55 & 49.83 \\
\textit{+~ExpeL} & 44.05 & 40.98 & 46.85 & 29.10 & \textbf{52.16} & 53.99 & 52.36 & 45.57 & 63.00 & 59.33 & 49.12 \\
\textit{+~GEPA} & 42.46 & 41.59 & 42.31 & 22.39 & 50.59 & 50.92 & 54.55 & 48.93 & 60.86 & 56.88 & 47.61 \\
\rowcolor{oursbg}\textbf{+ Ours} & \textbf{48.81} & \textbf{46.79} & 46.50 & \textbf{30.97} & 49.41 & \textbf{57.36} & 54.18 & 48.62 & 64.53 & 61.47 & \textbf{51.35} \\
\textit{Gain (pp)} & \textcolor{accent}{+2.78} & \textcolor{accent}{+1.83} & \textcolor{gray}{-2.10} & \textcolor{accent}{+4.48} & \textcolor{gray}{-1.57} & \textcolor{accent}{+3.99} & \textcolor{gray}{-2.18} & \textcolor{gray}{-0.61} & \textcolor{gray}{-0.92} & \textcolor{gray}{-0.61} & \textcolor{accent}{+0.51} \\
\bottomrule
\end{tabularx}%
\endgroup
\end{minipage}%
}
\end{table}

%% file: sections/appendix_case_studies.tex
\newcommand{\CasePanel}[3]{%
  \clearpage
  \thispagestyle{fancy}%
  \noindent\begin{center}%
    \includegraphics[page=#1,height=0.90\textheight,width=0.96\linewidth,keepaspectratio]{figures/case_study_panel.pdf}\par
    \vspace{3pt}%
    {\small\captionof{figure}{#2}\label{#3}\par}%
  \end{center}%
}
\newcommand{\CaseContinuation}[1]{%
  \clearpage
  \thispagestyle{fancy}%
  \noindent\begin{center}%
    \includegraphics[page=#1,height=0.93\textheight,width=0.96\linewidth,keepaspectratio]{figures/case_study_panel.pdf}\par
    \vspace{3pt}%
    {\small Figure~\ref{#1} (continued).\par}%
  \end{center}%
}

\begin{figure}[p]
\subsection{Qualitative Case Studies}
\label{app:case-study}
\centering
\includegraphics[page=1,height=0.86\textheight,width=0.96\linewidth,keepaspectratio]{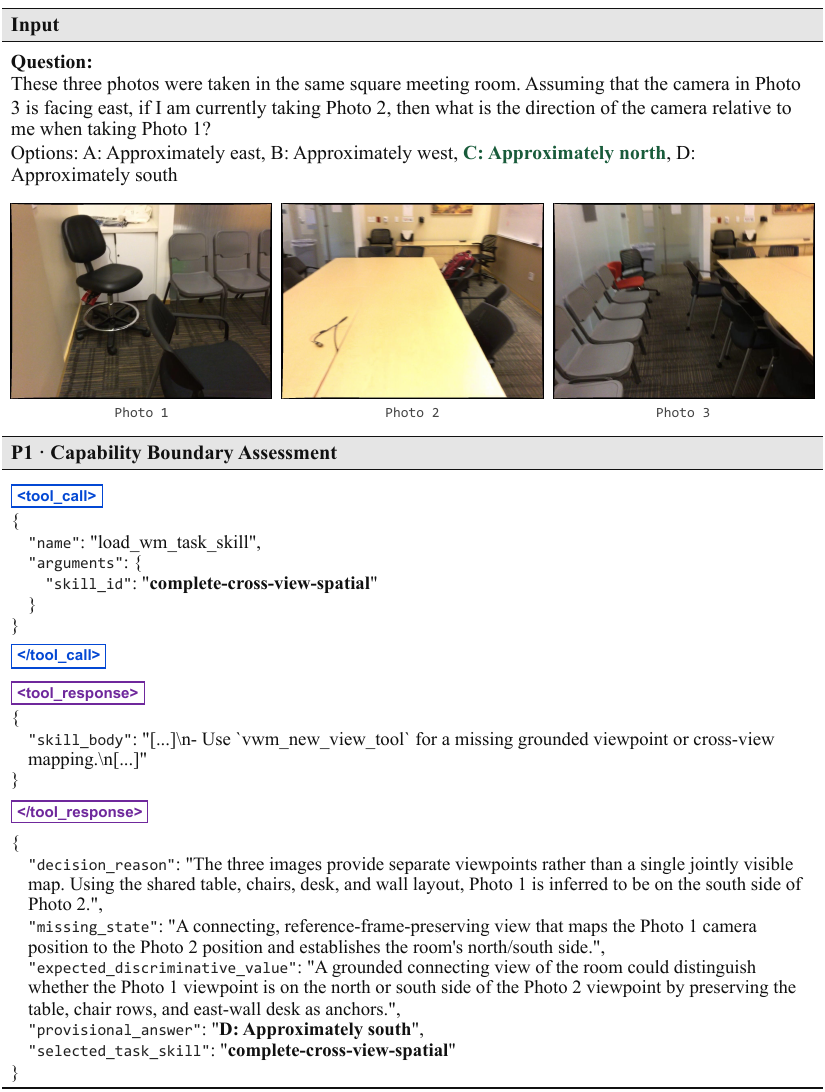}
\caption{Reasoning trajectory for \texttt{MMSI-Bench/MMSI-0071}.}
\label{fig:case-study-panel}
\end{figure}

\newcommand{\CaseContinuationOf}[2]{%
  \clearpage
  \thispagestyle{fancy}%
  \noindent\begin{center}%
    \includegraphics[page=#1,height=0.93\textheight,width=0.96\linewidth,keepaspectratio]{figures/case_study_panel.pdf}\par
    \vspace{3pt}%
    {\small Figure~\ref{#2} (continued).\par}%
  \end{center}%
}

\CaseContinuationOf{2}{fig:case-study-panel}
\CaseContinuationOf{3}{fig:case-study-panel}

\CasePanel{4}{Reasoning trajectory for \texttt{Spin-Bench/SPIN-2511}.}{fig:case-study-spin}
\CaseContinuationOf{5}{fig:case-study-spin}
\CaseContinuationOf{6}{fig:case-study-spin}

\CasePanel{7}{Reasoning trajectory for \texttt{SpatiaLab/SPATIALAB-590}.}{fig:case-study-removal}
\CaseContinuationOf{8}{fig:case-study-removal}
\CaseContinuationOf{9}{fig:case-study-removal}

\CasePanel{10}{Reasoning trajectory for \texttt{CausalSpatial/CS-COLL-6169db902026}.}{fig:case-study-motion}
\CaseContinuationOf{11}{fig:case-study-motion}
\CaseContinuationOf{12}{fig:case-study-motion}

\CasePanel{13}{Reasoning trajectory for \texttt{MMSI-Bench/MMSI-0821}.}{fig:case-study-verification}
\CaseContinuationOf{14}{fig:case-study-verification}
\CaseContinuationOf{15}{fig:case-study-verification}

\CasePanel{16}{Reasoning trajectory for \texttt{CausalSpatial/CS-COLL-6a01e625175f}.}{fig:case-study-intervention}
\CaseContinuationOf{17}{fig:case-study-intervention}
\CaseContinuationOf{18}{fig:case-study-intervention}